\def\year{2016}\relax
\documentclass[letterpaper]{article}
\usepackage{aaai19}
\usepackage{times}
\usepackage{helvet}
\usepackage{courier}
\usepackage{amssymb}
\usepackage{amsmath}
\usepackage{amsthm}
\usepackage{array}
\usepackage{algorithm}
\usepackage{algorithmic}
\usepackage[tight,footnotesize]{subfigure}
\usepackage{graphicx}
\usepackage{mathrsfs}
\usepackage{multirow}
\usepackage{mdwmath}
\usepackage{mdwtab}

\newcommand{\citet}[1]{\citeauthor{#1}~\shortcite{#1}}
\newcommand{\citep}{\cite}

\begin{document}
\title{Multi-Precision Quantized Neural Networks via Encoding \\ Decomposition of $\{-1,+1\}$}
\author{Qigong Sun, \,Fanhua Shang, \,Kang Yang, \,Xiufang Li, \,Yan Ren, \,Licheng Jiao\\
Key Laboratory of Intelligent Perception and Image Understanding of Ministry of Education,\\
International Research Center for Intelligent Perception and Computation,\\
Joint International Research Laboratory of Intelligent Perception and Computation,\\
School of Artificial Intelligence, Xidian University, Xi'an, Shaanxi Province, 710071, China}
\maketitle

\begin{abstract}
The training of deep neural networks (DNNs) requires intensive resources both for computation and for storage performance. Thus, DNNs cannot be efficiently applied to mobile phones and embedded devices, which seriously limits their applicability in industry applications. To address this issue, we propose a novel encoding scheme of using $\{-1,+1\}$ to decompose quantized neural networks (QNNs) into multi-branch binary networks, which can be efficiently implemented by bitwise operations (\emph{xnor} and \emph{bitcount}) to achieve model compression, computational acceleration and resource saving. Based on our method, users can easily achieve different encoding precisions arbitrarily according to their requirements and hardware resources.
The proposed mechanism is very suitable for the use of FPGA and ASIC in terms of data storage and computation, which provides a feasible idea for smart chips.
We validate the effectiveness of our method on both large-scale image classification tasks (e.g., ImageNet) and object detection tasks. In particular, our method with low-bit encoding can still achieve almost the same performance as its full-precision counterparts.
\end{abstract}

\section{Introduction}
Deep Neural Networks (DNNs) have been successfully applied in many fields, especially in image classification, object detection and natural language processing. Because of numerous parameters and complex model architectures, huge storage space and considerable power consumption are needed. Furthermore, with the rapid development of chip technology, especially GPU and TPU, the computing power has been greatly improved. In the rapid developing era of deep learning, researchers use multiple GPUs or computer clusters to contribute to the exploration of complex problems. Nevertheless, the energy consumption and limitation of computing resources are still significant factors in industrial applications, which are generally ignored in scientific research.
In other words, breathtaking results of many DNNs algorithms under the condition of applying GPUs lag behind the demand of industry. DNNs can hardly be applied in mobile phones and embedded devices (as typical industrial applications) directly due to their limited memory and calculation resources. Therefore, the compression and acceleration of networks are especially important in future development and commercial applications.

In recent years, many solutions have been proposed to improve the energy efficiency of hardware, achieve model compression or computational acceleration, such as network sparse and pruning \cite{Hassibi1993Second,Wen2016Learning,Tran2015Learning}, low-rank approximation \cite{denton2014exploiting,jaderberg2014speeding,tai2015convolutional},
architecture design \cite{Howard2017MobileNets,Sandler2018MobileNetV2,LuoThiNet}, model quantization \cite{hubara2017quantized,Rastegari2016XNOR,lin2017towards}, and so on.
Network sparse and pruning can dramatically reduce the redundant connections, and thus reduce the computational load in the inference process without large accuracy drop.
\citet{tai2015convolutional} used low-rank tensor decomposition to remove the redundancy in the kernels which can be as a generic tool for speeding up. Since there is some redundant information in the networks, the most direct approach of cutting down those information is to optimize the structure and yield small networks \cite{Ioffe2015Batch,Iandola2016SqueezeNet}.
For example, \citet{Howard2017MobileNets} proposed to use bitwise separable convolutions to build light networks for mobile applications. Most of those networks still utilize floating-point number representations (i.e., full-precision values). However, \citet{gupta2015deep} discussed that the representation of the full-precision weights and activations in networks is not necessary during the training of DNNs, and a nearly identical or slightly better accuracy rate may be obtained under lower-precision representation and calculation.

Since non-differentiable discrete functions are applied in QNNs generally, there obviously exists the gradient mismatch problem in training process. Therefore, the backpropagation algorithm cannot be directly used to train QNNs. Many scholars \cite{mishra2017apprentice,polino2018model,tang2017train,wang2018two} are devoted themselves to improving the performance (e.g., accuracy and compression ratio) of QNNs, but few researchers have studied their acceleration, which is an important reason for hindering industrial promotion. To the best of our knowledge, the accelerated method used in binarized neural networks (BNNs) \cite{Courbariaux2016Binarized} is the most efficient strategy at present.
This strategy uses bitwise operations (\emph{xnor} and \emph{bitcount}) to replace full-precision matrix multiplication, and results 58$\times$ faster and 32$\times$ memory saving in CPU \cite{Rastegari2016XNOR}. As discussed in \cite{Liang2017FP}, it has a higher acceleration ratio on FPGA, which can speed up to about 705$\times$ in the peak condition compared with CPU and is 70$\times$ faster than GPU. In particular, they quantized activation values and weights to bits and used bitwise logic operations to achieve extreme acceleration ratio in inference process, but they could suffer significant performance degradation. However, most models were proposed for a fixed precision, and cannot extend to other precision models. They may easily fall into local optimal solutions and suffer from slow convergence speed in training process.

The representation capability of binary parameters is insufficient for many practical applications, especially for large-scale image classification (e.g., ImageNet) and regression tasks.
In order to address various complex problems and take full advantage of bitwise operations, \citet{lin2017towards} used the linear combination of multiple binary parameters \{-1, +1\} to approximate full-precision weights and activations. Therefore, the complex full-precision matrix multiplication can be decomposed into some simpler operations. This is the first time to use binary networks for image classification on ImageNet. \citet{Guo2017Network} and \citet{Xu2018Alternating} used the same technique to accelerate the training of CNNs and RNNs. In addition, those methods not only increase the number of parameters many times, but also introduce a scale factor to transform the original problem into an NP-hard problem, which naturally makes the solution difficult and high complexity.

In order to bridge the gap between low-bit and full-precision and apply to many cases, we propose a novel encoding scheme of using $\{-1,+1\}$ to easily decompose trained QNNs into multi-branch binary networks. Therefore, the inference process can be efficiently implemented by bitwise operations (\emph{xnor} and \emph{bitcount}) to achieve model compression, computational acceleration and resource saving.
Thus, our encoding mechanism can improve the utilization of hardware resources, and achieve parameter compression and computation acceleration. In our experiments, we not only validate the performance of our method for image classification on CIFAR-10 and large-scale datasets, e.g., ImageNet, but also implement object detection tasks. The advantages of our method are shown as follows:
\begin{itemize}
  \item We can directly use the high-bit model parameters to initialize a low-bit model for faster training. Hence, our networks can be trained in a short time, and only dozens of times fine-tuning are needed to achieve the accuracies in our experiments. Of course, we can get better performance if we continue training the network. Thus, our multi-precision quantized networks can be easily popularized and applied to engineering practices.
  \item We propose a range of functions (called MBitEncoder) to decompose activations (for example, we can use $M$ functions to get the state $\{-1, +1\}$ of $M$ encoded bits), which are used for inference computation. Therefore, those decomposed bits can be directly used in network computation without other judgments and mapping calculations.
  \item After the process of decomposition, instead of storing all encoding bits in data types, e.g., char, int, float or double, the parameters can be individually stored by bit vectors. Thus, the smallest unit of data in electronic equipments can be reduced to 1-bit from 8-bit, 16-bit, 32-bit or 64-bit, which raised the utilization rate of resources and compression ratio of the model. Then the data can be encoded, calculated and stored in various encoding precisions.
\end{itemize}

\section{Related Work}
QNNs can effectively implement model compression, even to 32$\times$ memory saving. Many researchers are focusing on the following three classes of methods: quantification methods, the methods of optimization in training process and acceleration computation in inference process.

Quantification methods play a significant role in QNNs, and determine the state and distribution of weights and activation values.
\citet{gupta2015deep} used the notation of integer and fractional to denote a 16-bit fixed-point representation, and proposed a stochastic rounding method to quantify values. \citet{vanhoucke2011improving} used 8-bit quantization to convert weights into signed char and activation values into unsigned char, and all the values are integer. For multi-state quantification (8-bit to 2-bit), linear quantization is usually used in \cite{hubara2017quantized,wang2018two,zhuang2018towards}.
Besides, \citet{miyashita2016convolutional} proposed logarithmic quantization to represent data and used bitshift operation in log-domain to compute dot products.
For ternary weight networks \cite{Li2016Ternary}, the weights are quantized to $\{-\Delta^*,0,+\Delta^*\}$, where $\Delta^*=0.7\cdot E(|W|)$.
In \cite{Zhu2016Trained}, the positive and negative states are trained together with other parameters.
When the states are constrained to 1-bit, \citet{Courbariaux2016Binarized} applied the sign function to binarize weights and activation values \{-1, +1\}.
In \cite{Rastegari2016XNOR}, the authors also used $\{-\alpha^*, +\alpha^*\}$ to represent the binary states, where $\alpha^*=\frac{1}{n}\|W\|_{\emph{l}_1}$.

It is obvious that discrete functions, which are non-differentiable or have zero derivatives everywhere, need to quantize weights or activation values.
The traditional gradient descent method is unsuitable for the training of deep networks. Recently, there are many researchers devoting themselves to addressing this issue.
\citet{li2017training} divided optimization methods into two categories: quantizing pre-trained models with or without retraining \cite{lin2016fixed,zhou2017incremental,Li2016Ternary,lin2017towards} and directly training quantized networks \cite{Courbariaux2015BinaryConnect,Courbariaux2016Binarized,Rastegari2016XNOR,wang2018two}.
\cite{Courbariaux2016Binarized,hubara2017quantized} used the straight-through estimator (STE) in \cite{Bengio2013Estimating} to train networks.
STE uses the nonzero gradient to approximate the function gradient, which is not-differentiable or whose derivative is zero, and then applies the stochastic gradient descent (SGD) to update the parameters. \citet{mishra2017apprentice,polino2018model} applied knowledge distillation techniques, which use high-precision teacher network to guide low-precision student network to improve network performance. In addition, some networks as in \cite{Guo2017Network,lin2017towards} use the linear combination of binary values to approximate the full-precision
weights and activation values. They not only increase the number of parameters many times, but also introduce the scale factor to transform the original problem into an NP-hard problem, which naturally makes the solution difficult and high complexity. \citet{Xu2018Alternating} used the two valued search tree to optimize the scale factor and achieved better performance in the language model by using the quantized recurrent neural networks.

After quantizing, weights or activation values are represented in a low-bit form, which has the potential of acceleration computation and memory saving. Because the hardware implementation has a certain threshold, many scholars have avoided considering their engineering acceleration. This is also an important reason for hindering industrial promotion.
The most direct quantization is to convert floating-point parameters into their fixed-point (e.g., 16-bit, 8-bit), which can achieve hardware acceleration for fixed-point based computation \cite{gupta2015deep,vanhoucke2011improving}. When the weight is extremely quantized to the binary weight \{-1, +1\} as in \cite{Courbariaux2015BinaryConnect} or ternary weight \{-1, 0, +1\} as in \cite{Li2016Ternary}, the matrix multiplication can be transformed into full-precision matrix addition and subtraction to accelerate computation.
Especially when the weight and activation values are binarized, matrix multiplication operations can be transformed into highly efficient logical and bitcounting operations \cite{Courbariaux2016Binarized,Rastegari2016XNOR}. \citet{Guo2017Network,lin2017towards} used a series of linear combinations of \{-1, +1\} to approach the parameters of full-precision convolution model, and then converted floating point operations into multiple binary weight operations to achieve model compression and computation acceleration.

\section{Multi-Precision Quantized Neural Networks}
\label{gen_inst}
In this section, we use the multiplication of two vectors to introduce the novel encoding scheme of using $\{-1,+1\}$ to decompose QNNs into multi-branch binary networks. In each branch binary network, we use -1 and +1 as the basic elements to efficiently achieve model compression and forward inference acceleration for QNNs. Different from fixed-precision neural networks (e.g., binary, ternary), our method can yield multi-precision networks and make full use of the advantage of bitwise operations to accelerate QNNs.

\subsection{Model Decomposition}
As the basic computation in most neural network layers, matrix multiplication costs lots of resources and also is the most time consuming operation.
Modern computers store and process data in binary format, thus non-negative integers can be directly encoded by \{0, 1\}.
We propose a novel decomposition method to accelerate matrix multiplication as follows: Let $x\!=\![\mathrm{x}^1,\mathrm{x}^2,...,\mathrm{x}^N]^T$ and $w\!=\![\mathrm{w}^1,\mathrm{w}^2,...,\mathrm{w}^N]^T$ be two vectors of non-negative integers, where $\mathrm{x}^i,\mathrm{w}^i \in \{0,1,2,...\}$ for $i\!=\!1,2,...,N$. The dot product of those two vectors can be represented as follows:
\begin{eqnarray}
x^T\cdot w =\: [\mathrm{x}^1,\mathrm{x}^2,...,\mathrm{x}^N][\mathrm{w}^1,\mathrm{w}^2,...,\mathrm{w}^N]^T \\
=\,\sum_{n=1}^{N}\mathrm{x}^n\cdot \mathrm{w}^n. \quad\quad\quad\quad\quad\quad\quad\quad\ \
\end{eqnarray}

All of the above operations consist of $N$ multiplications and $(N-1)$ additions. Based on the above \{0, 1\} encoding scheme, the vector $x$ can be encoded to binary form using $M$ bits, i.e.,
\begin{eqnarray}\label{equ01}
x\!=\![\overbrace{\mathrm{c}^1_M\:\! \mathrm{c}^1_{M-1}...\:\!\mathrm{c}^1_1},\overbrace{\mathrm{c}^2_M\:\! \mathrm{c}^2_{M-1}...\:\!\mathrm{c}^2_1},...,\overbrace{\mathrm{c}^N_M \:\!\mathrm{c}^N_{M-1}...\:\!\mathrm{c}^N_1}]^{T}.
\end{eqnarray}
Then the right-hand side of (\ref{equ01}) can be converted into the following form:
\begin{eqnarray}
\begin{bmatrix}
 \mathrm{c}^1_M&  \mathrm{c}^2_M&  \cdots & \mathrm{c}^N_M\\
 \mathrm{c}^1_{M-1}&  \mathrm{c}^2_{M-1}& \cdots & \mathrm{c}^N_{M-1}\\
 \vdots &  \vdots& \cdots & \vdots\\
 \mathrm{c}^1_1&  \mathrm{c}^2_1& \cdots & \mathrm{c}^N_1
\end{bmatrix}
=\begin{bmatrix}
 c_M\\
c_{M-1}\\
 \vdots\\
 c_1
\end{bmatrix},
\end{eqnarray}
where
\vspace{-2mm}
\begin{eqnarray}
\mathrm{x}^j\!&=&\!\sum_{m=1}^{M}2^{m-1}\cdot  \mathrm{c}^j_m, \quad \mathrm{c}^j_m\in \{0,1\},\\
c_i\!&=&\![\mathrm{c}^1_i,\mathrm{c}^2_i,...,\mathrm{c}^N_i].
\end{eqnarray}
In such an encoding scheme, the number of represented states is not greater than $2^M$. In addition, we encode another vector $w$ with $K$-bit numbers in the same way.
Therefore, the dot product of the two vectors can be computed as follows:
\begin{eqnarray}
x^T\cdot w =\sum_{n=1}^{N}\mathrm{x}^n\cdot \mathrm{w}^n    \quad\quad\quad\quad\quad\quad\quad\quad\quad\quad \quad\quad\quad\\
=\sum_{n=1}^{N}\left(\sum_{m=1}^{M}2^{m-1}\cdot \mathrm{c}^n_m\right)\cdot \left(\sum_{k=1}^{K} 2^{k-1}\cdot \mathrm{d}^n_k\right)\\
=\sum_{m=1}^{M}\sum_{k=1}^{K}2^{m-1}\cdot 2^{k-1}\cdot c_m\cdot d_k^T. \quad\quad\quad\quad\quad
\end{eqnarray}

From the above formulas, the dot product is decomposed into $M\times K$ sub-operations, in which every element is 0 or 1.
Because of the restriction of encoding and without using the sign bit, the above representation can only be used to encode non-negative integers.
However, it¡¯s impossible to limit the weights and the values of the activation functions to non-negative integers.

\begin{figure*}
\begin{center}
\includegraphics[width=6.0in]{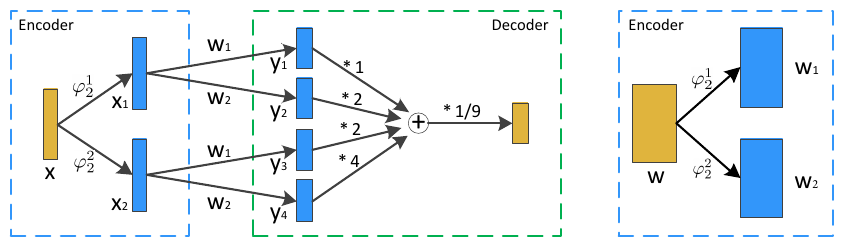}
\end{center}
\caption{Architecture of fully connected layer by 2-bit encoding. We use 2BitEncoder ($\varphi _{2}^{1}(x)$ and $\varphi _{2}^{2}(x)$) to encode
input data and weights in Encoder part and sum over those four results by fixing scale factors to achieve the final output in Decoder part.}
\end{figure*}

In order to extend encoding space to negative integer and reduce the computational complexity, we propose a new encoding scheme,
which uses \{-1, +1\} as the basic elements of our encoder rather than \{0, 1\}.
Except for the difference of basic elements, the encoding scheme is similar to the rules shown in the formula (5), and is formulated as follows:
\begin{eqnarray}
x^i=\sum_{m=1}^{M}2^{m-1}\cdot \mathrm{c}^i_m, \quad \mathrm{c}^i_m\in \{-1,1\}.
\end{eqnarray}
where $M$ denotes the number of encode bit, that can represent $2^M$ states. At this time, we can use multiple bitwise operations (\emph{xnor} and \emph{bitcount}) to effectively achieve the above vector multiplications. This operation mechanism is suitable for all vector/matrix multiplications.

In neural networks, matrix multiplication is the basic computation in both the fully connected and convolution layers. Based on the above decomposition mechanism of vector multiplication, we propose the following model decomposition method for quantized networks. We first use 2-bit encoding for fully connected layer as an example to introduce the mechanism of our model decomposition, the details are shown in Figure 1.
$x$ is the input data and $w$ is the weight matrix. Here, we suppose the bias does not exist. We define an "Encoder" that can be used in the 2BitEncoder function ($\varphi _{2}^{1}(\cdot)$ and $\varphi _{2}^{2}(\cdot)$), which will be described in the next section, to encode input data. For example, $x$ can be encoded by $x_1\in \{-1, +1\}^N$ and $x_2\in \{-1, +1\}^N$,
where $x_2$ represents high bit data and $x_1$ represents low bit data. These variables meet the following formula: $x=x_1 + 2x_2$.
In the same way, the weight $w$ can be converted into $w_1\in \{-1, +1\}^{M\times N}$ and $w_2\in \{-1, +1\}^{M\times N}$.
After cross multiplications, we get four intermediate variables \{$y_1, y_2, y_3, y_4$\}.
Each multiplication can be considered as a binarized fully connected layer, whose elements are -1 or +1.
This decomposition can result multi-branch layers, thus we call it as Multi-Branch Binary Networks (MBNs).
For instance, we decompose the 2-bit fully connection operation into four branches binary operations, which can be accelerated by bitwise operations, and then sum over those four results by fixing scale factors to achieve the final output. This operation mechanism can be suitable for all vector/matrix multiplications. In addition to fully connected layers, convolution and deconvolution layers are also suit for neural networks.

\subsection{M-bit Encoding Functions}
As an important part in neural networks, activation functions can enhance the nonlinear characterization of networks. In our proposed model decomposition method, encoding function plays a critical role and can encode input data to multi-bits (-1 or +1). Those numbers represent the encoding of input data. For some other QNNs, several quantization functions have been given. However, it is not clear that what's the affine mapping between quantized numbers and encode bits. In this part, a list of $M$-bit encoding functions are proposed to produce the element of each bit that follows the rules for encoding data.

\begin{table}[htbp]
\centering
\caption{Activation functions to limit input data to a fixed numerical range.}
\small
\begin{tabular}{c|c}
\hline
\hline
\!\!\!\! $Tanh(x)=\frac{e^x-e^{-x}}{e^x+e^{-x}}$ & $HTanh(x)= \left\{\begin{matrix}
+1, & x>1\\
x, & -1\leqslant x\leqslant 1\!\!\\
-1, & -1\leqslant x
\end{matrix}\right.
$ \\
\hline
\!\!\!\! $Sigmoid(x)=\frac{1}{1+e^{-x}}$ & $HReLU(x)= \left\{\begin{matrix}
+1, & x>1\\
x, & 0\leqslant x\leqslant 1\!\!\\
0, & x\leqslant 0
\end{matrix}\right.
$ \\
\hline
\hline
\end{tabular}
\end{table}

Before encoding, the data should be limited to a fixed numerical range. Table 1 lists four activation functions. $HTanh(\cdot)$ brings the range of input data to [-1, +1], and it consists with sign function to achieve binary encoding of weights and activations \cite{Courbariaux2016Binarized,Liang2017FP}. Since the convergence of SGD obtained by using $ReLU(\cdot)$ is faster than other activation functions, we propose a new activation function $HReLU(\cdot)$ that retains the linear characteristics in the specific range and limits the range of input data to [0, 1].
Different from general activation functions mentioned above, the output of our $M$-bit encoding function defined below should be $M$ numbers, which is -1 or +1. Those numbers represent the encoding of input data. Therefore, the dot product can be computed by the formula (9). In addition, at the above described experimental condition, when we use 2-bit to encode the data x and constrain to [-1, 1], there are 4 encoded states, as shown in Table 2. The affine mapping between quantized real numbers and their encoded states is given in the following table.

\begin{table}[htbp]
\centering
\caption{Quantized real numbers and their Encoded states.}
\centering
\renewcommand\arraystretch{1.5}
\begin{tabular}{c c c c c}
\hline
\hline
Quantized numbers & -1 & -1/3 & 1/3 & 1 \\
\hline
Encoded states & \{-1,-1\} & \{-1,1\} & \{1,-1\} & \{1,1\} \\
\hline
\hline
\end{tabular}
\end{table}

\begin{figure*}
\begin{center}
\includegraphics[width=6.0in]{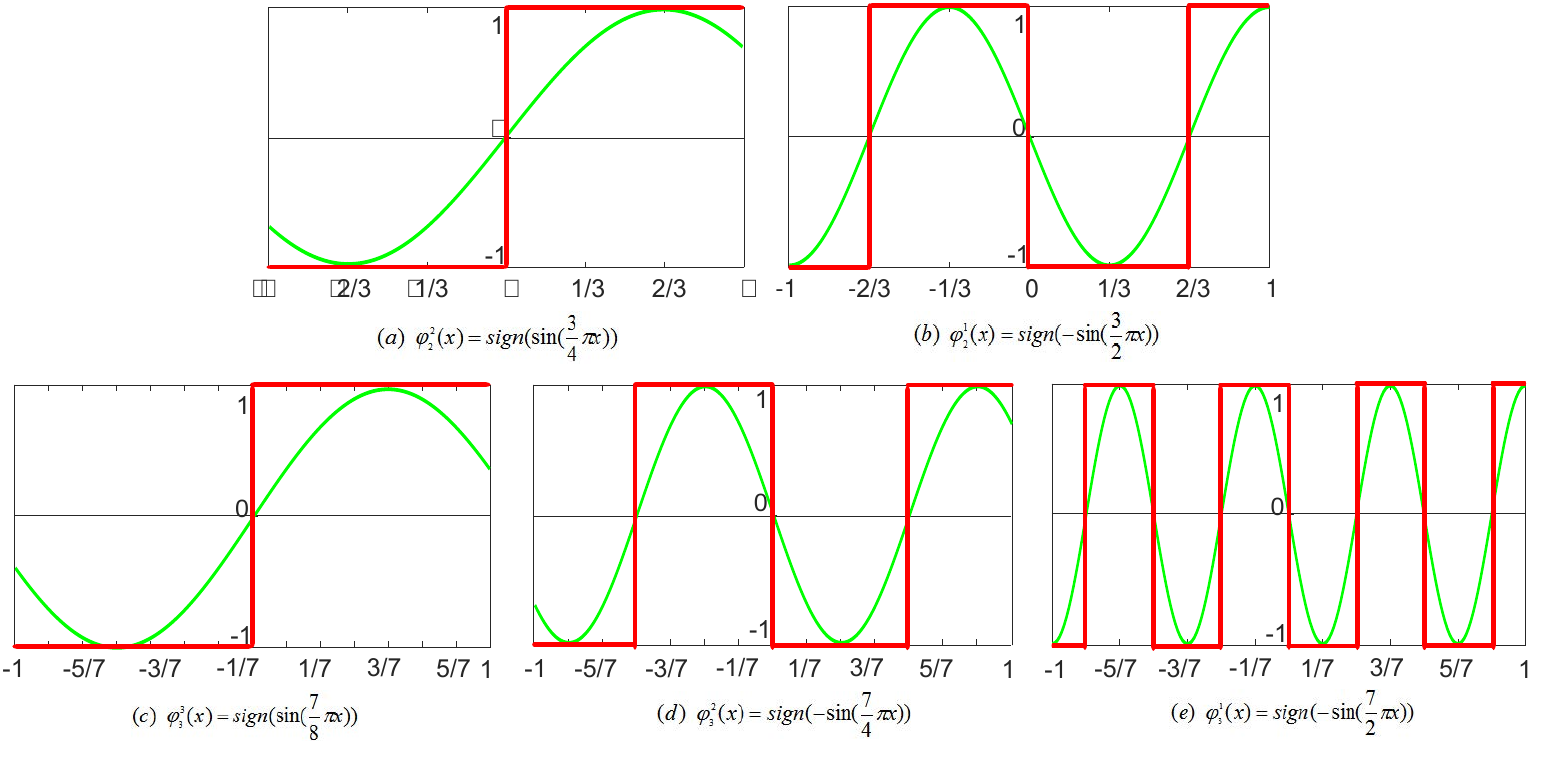}
\end{center}
\caption{Encoding functions. (a) and (b) denote the encoding functions of the second bit and the first bit of 2BitEncoder.
(c), (d) and (e) denote the encoding functions of the third bit, the second bit and the first bit of 3BitEncoder.}
\end{figure*}

From the above results, we can see that there is a linear factor $\alpha$ between quantized real numbers and encoded states (e.g., $\alpha$=3 for Table 2). When we use formula (9) to compute the multiplication of two encoded vectors, the value will be expanded $\alpha ^2$ times. Therefore, the result can multiply its scale factor to get the final result, shown as $1/9$ in Figure 1. Figure 2 shows the illustration of 2-bit and 3-bit encoding functions, we can see that those encoding functions are required periodic, and each function has different periods. Naturally, we apply trigonometric functions as the basic encoder functions, which are signed as red lines. After all, we use sign function to hard divide to -1 or +1. The mathematical expression can be formulated as follows:
\begin{eqnarray}
2BitEncoder(x)=\left\{\begin{matrix}
\varphi _{2}^{2}(x):sign(sin(\frac{3}{4}\pi \cdot x)), \quad \\
\varphi _{2}^{1}(x):sign(-sin(\frac{3}{2}\pi \cdot  x)), \quad\!\!\!\!\!
\end{matrix}\right.
\end{eqnarray}
where $\varphi _{2}^{1}(x)$ denotes the encoding function of the first bit ($x^i_1$) of 2BitEncoder, and $\varphi _{2}^{2}(x)$
represents the encoder function of the second bit ($x^i_2$) of 2BitEncoder. The periodicity is obviously different from others because it needs to denote more states.

\section{Networks Training}

QNNs face the problem that the derivative is not defined, thus traditional gradient optimization methods are not applicable. \citet{Courbariaux2016Binarized} presented the \emph{HTanh} function to binary quantize both weights and activations, and also defined the derivative to support back-propagation (BP)  training process.
They used the loss computed by binarized parameters to update full precision parameters.
Similarly, \citet{Rastegari2016XNOR} also proposed to update the weights with the help of the second parameters. \citet{Bengio2013Estimating} discussed that using STE to train network models containing discrete variables can obtain faster training speed and better performance.

\subsection{Multi-Branch Binary Networks Training}
Generated by the decomposition of QNNs, MBNs need to use $M$-bit encoding functions to get the elements of
each bit, which can be used by more efficient bitwise operations to replace arithmetic operations.
We take the 2-bit encoding as an example to describe the optimization method of MBNs. The sign function of the encoder makes it difficult to implement the BP process. Thus, we approximate the derivation of the encoder function with respect to $x$ as follows:
\begin{eqnarray}
\frac{\partial \varphi _{2}^{2}(x)}{\partial x}=\left\{\begin{matrix}
\frac{3}{4}\pi cos(\frac{3}{4}\pi x), & -1\leqslant x\leqslant 1 \\
0 & \textup{otherwise},
\end{matrix}\right.
\end{eqnarray}
\begin{eqnarray}
 \frac{\partial \varphi _{2}^{1}(x)}{\partial x}=\left\{\begin{matrix}
-\frac{3}{2}\pi cos(\frac{3}{2}\pi x), & -1\leqslant x\leqslant 1 \\
0 & \textup{otherwise}.
\end{matrix}\right.
\end{eqnarray}

Besides activations, all weights of networks also need to be quantized to binary values.
We retain the real-valued weight $w$ and binarized weight $w_b$ in the training process, and apply $w_b$ to compute loss and gradient, which is used to update $w$.
$w$ is constrained between -1 to +1 to avoid excessive growth. Different from weights, the binary function for $w$ is not needed for the encoding function and directly defined as follows:
\begin{eqnarray}
Binarize(x)=sign(HTanh(x)).
\end{eqnarray}
For this function, we have defined the gradient function of each component to constrain the search space. That is, the input of sign function can be constrained to [-1,+1] by $HTanh(x)$, and it can also speed up the convergence. The parameters of the whole network are updated by \emph{Adam} \cite{kingma2014adam} in the condition for differentiability.

\subsection{Quantized Networks Training}
The above training scheme is proposed to optimize binary networks, which can be converted into multi-state networks.
However, this converter can produce many times more parameters than the original network.
If we optimize the binarized network, it may easily fall into local optimal solutions and face slow convergence speed.
Based on the affine mapping between quantized numbers and fixed-point integers, we can directly optimize the quantized network
and then use multi-branch binary operations in inference process.
There are two quantization schemes usually applied in QNNs \cite{Courbariaux2016Binarized,Zhou2016DoReFa,miyashita2016convolutional}, named linear quantization and logarithmic quantization.
Due to the requirement of our encoding mechanism, linear quantization is used to quantize networks, and is defined as follows:
\begin{eqnarray}
q_k(x)=2\left(\frac{<(2^k-1)(\frac{x+1}{2})>}{2^k-1}-\frac{1}{2}\right),
\end{eqnarray}
where $<\!\cdot\!>$ denotes the rounding operation, which can quantize a real number $x\!\in\! [-1,+1]$ to a certainty state.
We call it a hard ladder function, which can segment input codomain to multi-states.
Table 2 lists the four states that quantized by formula (15).
However, the derivative of this function is almost zero everywhere, it cannot be used in training process. Inspired by STE, we use the same technique to speed up computing process and yield better performance. We use the loss computed by quantized parameters to update full precision parameters.
Note that for our encoding scheme with low-precision quantization (e.g., binary), we use \emph{Adam} to train our model, otherwise stochastic gradient descent is used.

\section{Experiments}

Many scholars are devoted to improving the performance (e.g., accuracy and compression ratio) of QNNs, while very few researchers have studied their engineering acceleration, which is an important reason for hindering industrial promotion. Therefore, we mainly focus on an acceleration method, which is especially suitable for engineering applications. In this section, we compare the performance of our method with BWN \cite{Courbariaux2015BinaryConnect}, BNN \cite{Courbariaux2016Binarized}, XNOR-Net \cite{Rastegari2016XNOR}, TWN \cite{Li2016Ternary}, and ABC-Net \cite{lin2017towards} for image classification tasks on CIFAR-10 and ImageNet, and object detection tasks on Pascal VOC2007/2012 datasets.

\subsection{Image Classification}

\begin{table*}[htbp]
\centering
\caption{Classification accuracies of Lenet on CIFAR-10 and ResNet-18 on ImageNet.}

\renewcommand\arraystretch{1.2}
\centering
\begin{tabular}{c c c c}
\hline
\hline
Method & CIFAR-10 & ImageNet(Top-1) & ImageNet(Top-5) \\
\hline
BWN \cite{Courbariaux2015BinaryConnect}& 90.10\% & 60.80\% & 83.00\%\\
\hline
BNN \cite{Courbariaux2016Binarized}& 88.60\% & 42.20\% & 67.10\% \\
\hline
XNOR-Net \cite{Rastegari2016XNOR}& - & 51.20\% & 73.20\% \\
\hline
TWN \cite{Li2016Ternary}& 92.56\% & 61.80\% & 84.20\% \\
\hline
ABC-Net[5-bit] \cite{lin2017towards}& - & 65.00\% & 85.90\% \\
\hline
Full-Precision & 91.40\% & 68.60\% & 88.70\% \\
\hline
\hline
\multicolumn{4}{c}{Encoded activations and weights }\\
\hline
\hline
MBN[M=K=1] & 90.39\% & 47.10\% & 71.70\%\\
\hline
MBN[M=K=2]& 91.06\% & 56.30\% & 79.48\% \\
\hline
MBN[M=K=3]& 91.27\% & 58.69\% & 81.84\% \\
\hline
MBN[M=K=4]& 91.15\% & 59.57\% & 82.35\% \\
\hline
MBN[M=K=5] & 90.92\% & 65.09\% & 86.42\%\\
\hline
MBN[M=K=6]& 91.01\% & 67.04\% & 87.69\% \\
\hline
MBN[M=K=7]& 90.20\% & 68.37\% & 88.47\% \\
\hline
MBN[M=K=8]& 90.43\% & 68.63\% & 88.70\% \\
\hline
\hline
\end{tabular}
\end{table*}

\begin{table*}[htbp]
\centering
\caption{Comparison with different encoding bits for object detection.}

\renewcommand\arraystretch{1.5}
\centering
\begin{tabular}{c| c| c| c| c}
\hline
\hline
Method & Full-Precision & MBN[M=K=8] & MBN[M=K=6] & MBN[M=K=5]\\
\hline
mAP & 0.6392 & 0.6351 & 0.6131 & 0.5423\\
\hline
\hline
\end{tabular}
\end{table*}

\textbf{CIFAR-10:} CIFAR-10 is an image classification benchmark dataset, which has 50000 training images and 10000 testing images.
All the images are 32 $\times$ 32 color images representing airplanes, automobiles, birds, cats, deer, dogs, frogs, horses, ships and trucks.

We validated our method by different bit encoding schemes, in which activations and weights are equally treated, that is, both of them use the same bit-encoding. Table 3 lists the results of our method and several state-of-the-art models mentioned above. Here we use the same network architecture as in \cite{Courbariaux2015BinaryConnect,Courbariaux2016Binarized} except for the encoding functions. We use $\emph{HTanh}(\cdot)$ as the activation function and employ \emph{Adam} to optimize all parameters of the network. From all the results, we can see that the representation capabilities of 1-bit and 2-bit are completely enough for small-scale datasets, e.g., CIFAR-10. Our method with low-precision encoding achieves nearly the same classification accuracy as high precision and full-precision models, while we can attain $\sim\!16\times$ memory saving compared with its full-precision counterpart. When activations and weights are constrained to 1-bit, our network structure is similar to BNN \cite{Courbariaux2016Binarized}, and our method yields even better accuracy mainly because of our proposed encoding functions.

\textbf{ImageNet:} We further examined the performance of our method with different bit encoders on the ImageNet ILSVRC-2012 dataset \cite{Russakovsky2015ImageNet}. This dataset consists of 1K categories images, and has over 1.2M images in the training set and 50K images in the validation set. We use Top-1 and Top-5 accuracies to report the classification performance. For large-scale training sets (e.g., ImageNet), it usually costs plenty of time and requires sufficient computing resources for classical full-precision models.
It will be more hard to train quantized networks, thus the initialization of parameter values is particularly important.
In this paper, we present $\emph{HReLU}(\cdot)$ as the activation function to constraint activations.
In particular, the full-precision model parameters activated by $\emph{ReLU}(\cdot)$ can be directly used as initialization parameters for our 8-bit quantized network. After a little number of fine-tuning, 8-bit quantized networks can be well-trained. Similarly, we use the 8-bit model parameters as the initialization parameters to train 7-bit quantized networks, and so on.
There has a special case, if we use $\emph{HReLU}(\cdot)$ and 1BitEncoder function to encode activations, all the activations will be constrained to +1. Here, we use $\emph{HTanh}(\cdot)$ as the activation function for 1-bit encoding. Note that we use \emph{SGD} to optimize parameters when encoding bit is not less than 3, and the learning rate is set to 0.1.
When the encode bit is 1 or 2, the convergent speed of \emph{Adam} is faster than \emph{SGD}, as discussed in \cite{Courbariaux2016Binarized,Rastegari2016XNOR}.

Table 3 lists the performance (e.g., accuracy, speedup ratio, memory saving ratio) of our method and several typical models mentioned above.
Those results show that our method with 1-bit encoding performs much better than BNN \cite{Courbariaux2016Binarized}. Similarly, our method with 5-bit encoding significantly outperforms ABC-Net[5-bit] \cite{lin2017towards}. Moreover, our networks can be trained in such a short time, and to achieve the accuracies in our experiments only needs dozens of times fine-tuning. Of course, if we continue training the network, we can get better performance.
Different from BWN and TWN, whose weights are only quantized rather than activation values, our method quantifies both weights and activation values simultaneously. Although BWN and TWN can obtain little higher accuracies than our method with 1-bit quantization model, our method obtains more speedup, and the speedup ratio of existing methods such as BWN and TWN is limited to $\sim \!2\times$. Due to limited and fixed expression ability, existing methods (such as BWN, TWN, BNN, XNOR-Net) can not satisfy higher precision requirements. In particular, our method can provide 64 available encoding choices, and hence our encoded networks with different encoding precisions have different speedup ratios, memory requirements and experimental precisions.

\subsection{Object Detection}
We also use the trained ResNet-18 with the Single Shot MultiBox Detector (SSD) framework  \cite{liu2016ssd} to validate object detection, in which the coordinate regression task coexists with classification tasks.
The normally regression task has higher requirement on value precision, therefore the application of object detection presents a new challenge for QNNs.

In this experiment, our model is trained on the VOC2007 and VOC2012 train/val set, and tested on the VOC2007 test set. ResNet-18 with the SSD framework  \cite{liu2016ssd} is used as the basic network. Here we use the trained model parameters in ImageNet classification to initialize SSD network parameters, after dozens of times fine-tuning the results are listed in Table 4. We use Mean Average Precision (mAP) as the criterion to evaluate the performance of our model.
It is clear that our method with 8-bit encoding scheme can yield very similar performance as its full-precision counterpart. When we use 6-bit to encode parameters, the evaluation index dropped by 0.0261. If the number of encode bits is constrained to 5, the performance of this task has visibly deteriorated, while our method can achieve $\sim\!5\times$ memory saving.

As the attempt in object detection tasks, our method yields good performance on the SSD framework. Similarly, it can be applied to other frameworks, e.g., R-CNN \cite{girshick2014rich}, Fast R-CNN \cite{ren2015faster}, SPP-Net \cite{he2014spatial} and YOLO \cite{redmon2016you}.

\section{Discussion and Conclusion}

\subsection{\{0, 1\} Encoding and \{-1, +1\} Encoding}

As described in \cite{Zhou2016DoReFa}, there exists an affine mapping between quantized numbers and fixed-point integers.
The quantized numbers are usually restricted to the closed interval [-1, +1]. For example, the mapping is formulated as follows:
\begin{eqnarray}
\mathrm{x}^q=\frac{2}{2^M-1}\mathrm{x}^{\{0,1\}}-1,
\end{eqnarray}
where $\mathrm{x}^q$ denotes a quantized number and $\mathrm{x}^{\{0,1\}}$ denotes the fixed-point integer encoded by 0 and 1.
We use a $K$-bit fixed-point integer to represent a quantized number $\mathrm{w}^q$. The product can be formulated as follows:
\begin{eqnarray}\label{equ09}
\mathrm{x}^q\cdot \mathrm{w}^q=\frac{4}{(2^M-1)(2^K-1)}\mathrm{x}^{\{0,1\}}\cdot \mathrm{w}^{\{0,1\}}-  \nonumber \\
\frac{2}{2^M-1}\mathrm{x}^{\{0,1\}}-\frac{2}{2^K-1}\mathrm{w}^{\{0,1\}}+1.
\end{eqnarray}
The right-hand side of (\ref{equ09}) is a polynomial, which has four terms. And each term has its own scaling factor.
The computation of $\mathrm{x}^{\{0,1\}}\cdot \mathrm{w}^{\{0,1\}}$ can be accelerated by bitwise operations, however, the polynomial and scaling factor will
increase the computational complexity.

For our proposed quantized binary encoding scheme (i.e., $\{-1, +1\}$), the product of two numbers is defined as
\begin{eqnarray}
\mathrm{x}^q\cdot \mathrm{w}^q=\frac{1}{(2^M-1)(2^K-1)}\mathrm{x}^{\{-1,1\}}\cdot \mathrm{w}^{\{-1,1\}},
\end{eqnarray}
where $\mathrm{x}^{\{-1,1\}}$ and $\mathrm{w}^{\{-1,1\}}$ denote the fixed-point integers encoded by -1 and 1.
Obviously, compared with the above encoding of $\{0,1\}$, the product can be more efficiently calculated by using our proposed encoding scheme.

\subsection{Linear Approximation and Quantization}
As described in \cite{lin2017towards,Guo2017Network,Xu2018Alternating}, the weight $\mathrm{w}$ can be
approximated by the linear combination of $K$ binary subitems \{$\mathrm{w}_1, \mathrm{w}_2,..., \mathrm{w}_K$\} and $\mathrm{w}_i\in \{-1, +1\}^N$,
which can replace arithmetic operations with more efficient bitwise operations. In order to obtain the combination, we need to solve the following problem
\begin{eqnarray}
\min_{\{\alpha_i,\mathrm{w}_i \}_{i=1}^K}\left \| \mathrm{w}-\sum_{i=1}^{K}\alpha_i \mathrm{w}_i\right \|^2,  \;\mathrm{w}\in \mathbb{R}^N.
\end{eqnarray}
When this approximation is used in neural networks, $\mathrm{w}_i$ can be considered as model weights.
However, the scale factor $\alpha_i$ is introduced in this approximation, and such a scheme also expands the parameters $K$ times.
Therefore, this approximation can convert the original model to a complicated binary network, which is hard to train \cite{li2017training} and easily falls into local optimal solutions.

For our method, we use the quantized parameters $\mathrm{w}^q$ to approximate $\mathrm{w}$ as follows:
\begin{eqnarray}
\mathrm{w}\approx \frac{1}{2^K-1}\mathrm{w}^q,\;\mathrm{w}\in [-1,1]^N,
\end{eqnarray}
where $\mathrm{w}^q$ is a positive or negative odd number, and its absolute value is not larger than $2^K\!-\!1$. Different from the above linear approximation, our method can achieve the quantized weights, and directly get the corresponding encoding elements. Thus, our networks can be more efficiently trained via our quantization scheme than the linear approximation.

\subsection{Conclusions}
In this paper, we proposed a novel encoding scheme of using \{-1, +1\} to decompose QNNs into multi-branch binary networks, in which we used bitwise operations (\emph{xnor} and \emph{bitcount}) to achieve model compression, computational acceleration and resource saving. In particular, we can use the high-bit model parameters to initialize a low-bit model and achieve good results in various applications. Thus, users can easily achieve different encoding precisions arbitrarily according to their requirements (e.g., accuracy and speed) and hardware resources (e.g., memory). This special data storage and calculation mechanism can yield great performance in FPGA and ASIC, and thus our mechanism is a feasible idea for smart chips. Future works will focus on improving the hardware implementation and chip technology, and exploring some ways to automatically select proper bits for various network architectures (e.g., VGG and ResNet).

\section{Acknowledgments}

This work was partially supported by the State Key Program of National Natural Science of China (No.\ 61836009), Project supported the Foundation for Innovative Research Groups of the National Natural Science Foundation of China (No.\ 61621005), the National Natural Science Foundation of China (Nos.\ U1701267, 61871310, 61573267, 61502369, 61876220, 61876221, 61473215, and 61571342), the Fund for Foreign Scholars in University Research and Teaching Programs (the 111 Project) (No.\ B07048), the Major Research Plan of the National Natural Science Foundation of China (Nos.\ 91438201 and 91438103), the Program for Cheung Kong Scholars and Innovative Research Team in University (No.\ IRT\_15R53), and the Science Foundation of Xidian University (No.\ 10251180018). Fanhua Shang is the corresponding author.

\bibliographystyle{aaai}

\end{document}